\documentclass[conference]{IEEEtran}
\IEEEoverridecommandlockouts

\usepackage{cite}
\usepackage{amsmath,amssymb,amsfonts}
\usepackage{algorithmic}
\usepackage{graphicx}
\usepackage{textcomp}
\usepackage{xcolor}

\usepackage{subfig}
\usepackage{amsmath}
\usepackage{amssymb}
\usepackage{hyperref}
\usepackage{gensymb}
\hypersetup{
    colorlinks=true,
    linkcolor=blue,
    citecolor=blue,
    filecolor=magenta,      
    urlcolor=blue,
    pdfpagemode=FullScreen,
    }
\usepackage{booktabs}
\usepackage{multirow}
\usepackage{multicol}
\usepackage{algorithmic}

\usepackage{algorithm2e}

\def\BibTeX{{\rm B\kern-.05em{\sc i\kern-.025em b}\kern-.08em
    T\kern-.1667em\lower.7ex\hbox{E}\kern-.125emX}}
\begin{document}

\title{Uncertainty-Aware and Decoder-Aligned Learning for Video Summarization\\
\thanks{This work has been accepted for presentation at the 2026 International Joint Conference on Neural Networks (IJCNN 2026).}
}

\author{\IEEEauthorblockN{1\textsuperscript{st} Omer Tariq}
\IEEEauthorblockA{\textit{Perception AI} \\
\textit{Neubility Inc.}\\
Seoul, South Korea \\
omer.tariq@neubility.co.kr}
\and
\IEEEauthorblockN{2\textsuperscript{nd} Syed Muhammad Raza}
\IEEEauthorblockA{\textit{Perception AI} \\
\textit{Neubility Inc.}\\
Seoul, South Korea \\
raza@neubility.co.kr}
\and
\IEEEauthorblockN{3\textsuperscript{rd} Jeongbae Son}
\IEEEauthorblockA{\textit{Perception AI} \\
\textit{Neubility Inc.}\\
Seoul, South Korea \\
thswjdqo92@neubility.co.kr}
}

\maketitle

\begin{abstract}
Video summarization aims to produce a compact representation of a long video by selecting a subset of temporally important segments that best reflect human preferences. This task is inherently difficult due to strong annotation subjectivity and the reliance on discrete decoding procedures, such as temporal segmentation and knapsack-based selection, during evaluation. Most existing approaches either learn deterministic importance scores that overlook these characteristics or adopt complex generative models that increase training and inference cost.
In this paper, we propose VASTSum, an \emph{uncertainty-aware and decoder-aligned} learning framework for video summarization that addresses both challenges within a single-pass model. The proposed method predicts probabilistic frame-level importance scores using a variational formulation, enabling explicit modeling of uncertainty arising from multi-annotator supervision. To account for subjectivity, particularly under binary annotations, we employ a supervision strategy that encourages alignment with plausible human annotation modes rather than enforcing a single consensus target. Furthermore, we introduce a decoder-aligned regularization that promotes stability of knapsack-based summary selection, reducing sensitivity to small perturbations in predicted scores.
We evaluate the proposed framework on the SumMe and TVSum benchmarks using standard rank-based metrics. Experimental results show consistent and competitive Kendall and Spearman correlations across multiple data splits, demonstrating improved robustness under annotation disagreement while maintaining efficient single-forward inference. These results indicate that explicitly modeling uncertainty and aligning learning objectives with the decoding stage provide a principled alternative to both deterministic and diffusion-based video summarization methods.
\end{abstract}


\section{Introduction}
\label{sec:intro}

Recently, short-form videos have become a dominant format on online platforms, reflecting a broader shift toward rapid consumption of visual content. Video summarization addresses this demand by selecting a short sequence of temporally coherent segments that preserves the essential content of a long video under a strict length budget. However, what constitutes an ``essential'' summary is inherently subjective, and standard benchmarks therefore provide multiple human annotations per video, notably SumMe~\cite{SumMe} and TVSum~\cite{TVSum}.

In supervised settings, a typical pipeline predicts frame-level importance scores and then produces a binary summary through a discrete decoding stage. Concretely, predicted scores are aggregated into keyshots via temporal segmentation and a budget-constrained subset is selected using a knapsack optimizer, commonly following the kernel temporal segmentation (KTS) protocols~\cite{KTS}. Because this decoding is discrete, small perturbations in predicted scores can change the selected keyshots, making evaluation sensitive to score calibration and local ranking decisions.

Most prior methods nevertheless learn a deterministic regressor toward an averaged importance target, collapsing multi-annotator disagreement into a single supervision signal and potentially obscuring distinct but valid annotation modes~\cite{VASNet,DSNet,son2024csta}. Recent generative formulations instead aim to model a distribution of summaries, for example via diffusion-based conditional generation, but iterative inference increases computational cost and can complicate deployment~\cite{kim2025summdiff,ho2020denoising}. Moreover, both deterministic and generative approaches are typically optimized with continuous objectives, whereas reported performance is determined after segmentation and knapsack selection, which introduces a training and evaluation mismatch.

To address these challenges, we propose VASTSum, an uncertainty-aware and decoder-aligned learning framework that preserves single-pass efficiency while explicitly accounting for subjective supervision and discrete decoding. We parameterize frame importance as a predictive distribution using a variational formulation, enabling uncertainty to reflect annotator disagreement and ambiguous content. We further optimize against plausible annotation modes rather than enforcing a single collapsed consensus target, and we introduce a decoder-aligned regularization that promotes stability of knapsack-based keyshot selection under small score perturbations. Experiments on SumMe and TVSum demonstrate consistent improvements in rank correlation across standard splits, supporting uncertainty-aware and decoder-aligned learning as a practical alternative to both deterministic attention models~\cite{son2024csta} and diffusion-based generators~\cite{kim2025summdiff}.

We introduce an uncertainty-aware and decoder-aligned framework that bridges the gap between deterministic attention scoring, exemplified by CSTA~\cite{son2024csta}, and diffusion-based generative summarization, exemplified by SummDiff~\cite{kim2025summdiff}. Our approach retains single-pass efficiency while explicitly addressing annotation subjectivity and decoding sensitivity.

\begin{enumerate}
  \item We propose VASTSum, a unified uncertainty-aware variational formulation for frame-importance prediction that outputs both a mean estimate and an uncertainty signal, explicitly capturing ambiguity induced by multi-annotator supervision.

  \item We train from plausible annotator-consistent modes rather than collapsing subjective labels into a single averaged target, reducing over-smoothed supervision while preserving distinct summarization preferences.

  \item We introduce a decoder-aligned stability regularization that promotes stable keyshot selection under the standard segmentation and knapsack decoding pipeline, mitigating training--evaluation mismatch and improving robustness.
\end{enumerate}

Section~\ref{sec:related} reviews related work on supervised video summarization, covering deterministic attention-based scorers and recent generative formulations. Section~\ref{sec:problem} formalizes the task and the segmentation-plus-knapsack decoding protocol. Section~\ref{sec:method} presents VASTSum, including the hierarchical segment-context scorer, the uncertainty-aware variational importance head, the mode-aligned supervision strategy, and the decoder-aligned stability regularization.
Section~\ref{subsec:exp_settings} describes the experimental setup, including datasets, evaluation protocol, and implementation details. Section~\ref{subsec:results} presents quantitative results and discussion, followed by Section~\ref{sec:ablation}, which provides ablation studies analyzing key components of the proposed method. Finally, Section~\ref{sec:conclusion} concludes the paper and discusses limitations and future directions.

\section{Related Work}
\label{sec:related}

\noindent\textbf{Supervised video summarization.}
Most supervised pipelines predict frame-level importance scores and then apply a discrete decoding stage, typically change-point segmentation followed by budgeted 0/1 knapsack selection.
A large body of work learns deterministic scorers using temporal models and attention to capture local and global dependencies (e.g., VASNet~\cite{VASNet}, PGL-SUM~\cite{pglsum}), along with detect-to-summarize and joint learning variants (e.g., DSNet~\cite{DSNet}, iPTNet~\cite{iPTNet}).
Recent multimodal methods further incorporate language or cross-modal alignment signals to improve semantic relevance (e.g., CLIP-It~\cite{CLIPIt}, A2Summ~\cite{A2Summ}).
While effective, these approaches commonly collapse multi-annotator subjectivity into a single target and optimize continuous losses, even though evaluation depends on segmentation and knapsack decoding, which can be sensitive to small score perturbations.
CSTA~\cite{son2024csta} improves efficiency and temporal modeling with CNN-based spatiotemporal attention, but remains a deterministic scoring framework.

\vspace{0.1cm}
\noindent\textbf{Generative diffusion-based summarization.}
Diffusion models~\cite{ho2020denoising} have recently been adapted to generate summary distributions, with SummDiff~\cite{kim2025summdiff} modeling multiple plausible summaries to reflect diverse human preferences.
However, diffusion-based inference typically requires iterative denoising steps, increasing computational cost and complicating deployment when single-pass efficiency is desired.
In contrast to both deterministic scorers and iterative diffusion generators, our method targets subjective supervision and discrete decoding directly within a single-forward framework by learning uncertainty-aware importance scores and aligning training with the segmentation-plus-knapsack selection stage.


\begin{figure*}[h!]
    \centering
    \includegraphics[width=\textwidth]{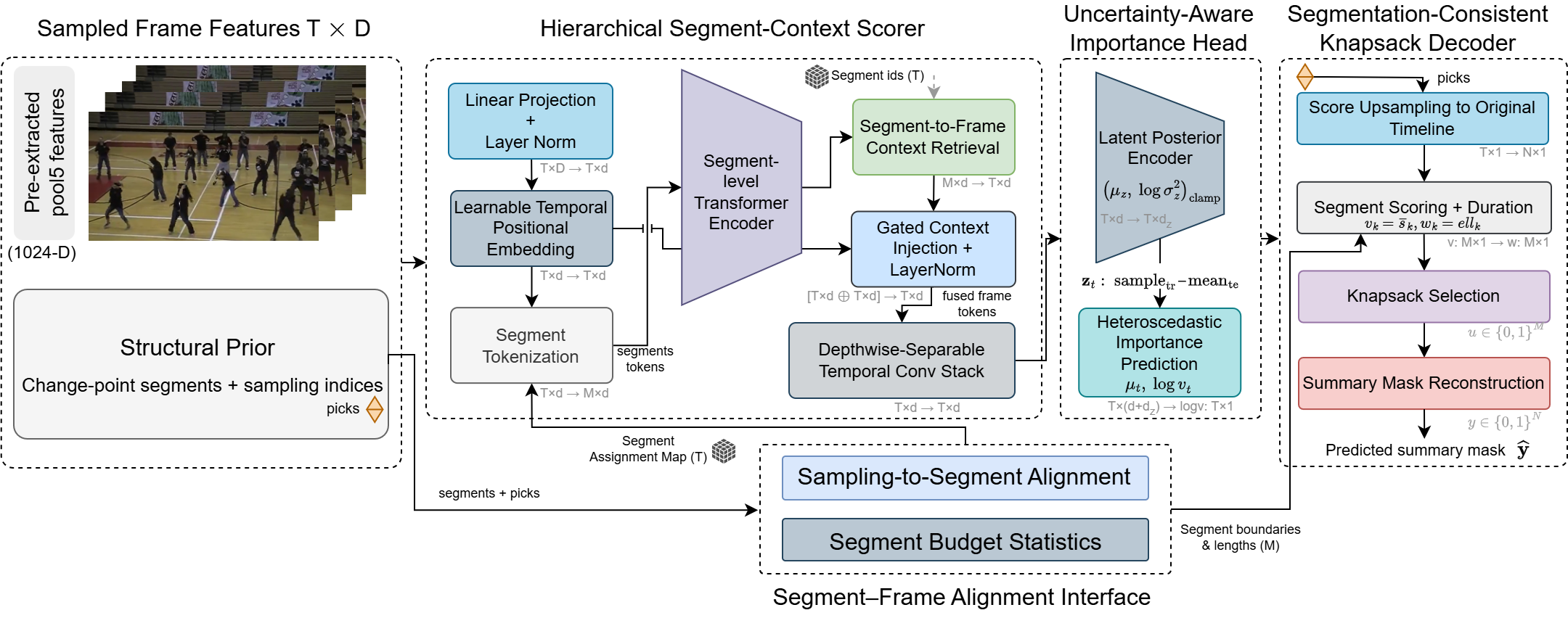}
    \caption{Overview of VASTSum: Proposed segmentation-consistent video summarization pipeline. Sampled \texttt{pool5} frame features are linearly projected and augmented with learnable temporal positional embeddings. Change-point segmentation provides a structural prior by assigning sampled timesteps to segment IDs and estimating segment boundaries/lengths. A hierarchical segment-context scorer aggregates frames into segment tokens, encodes global segment relations via a segment-level Transformer, propagates segment context back to frames with gated injection, and refines temporal representations using a lightweight depthwise-separable temporal convolution stack. A variational frame-importance head predicts a per-frame latent posterior and heteroscedastic importance parameters, sampling during training and using the posterior mean at inference for deterministic scoring. Predicted scores are upsampled to the full timeline using the sampling indices, and a segmentation-consistent 0--1 knapsack decoder selects segments under a length budget to output the binary summary mask.}
    
    \label{fig:pipeline}
\end{figure*}

\section{Problem Formulation}
\label{sec:problem}

Given an untrimmed video, we aim to produce a compact summary that preserves the essential content under a strict length budget. We follow the standard supervised setting where the input is a sequence of pre-extracted frame descriptors sampled from the original timeline.
Let
\begin{equation}
\mathbf{X}=\{\mathbf{x}_t\in\mathbb{R}^{D}\}_{t=1}^{T}
\end{equation}
denote the sampled feature sequence, where $D=1024$ in our implementation and $T$ is the number of sampled timesteps.
The sampled indices
\begin{equation}
\mathbf{p}=\{p_t\}_{t=1}^{T}, \qquad 0 \le p_1 < \cdots < p_T < N,
\end{equation}
map each sampled feature to its location on the original timeline of $N$ frames.
Benchmarks provide multiple human annotations per video.
Let $a\in\{1,\dots,U\}$ index annotators.
TVSum provides continuous frame-importance signals $s^{(a)}\in[0,1]^T$ aligned to the sampled timeline, while SumMe provides binary user summaries on the original timeline; in SumMe, we evaluate each annotator's inclusion signal at the sampled indices, yielding $y^{(a)}\in\{0,1\}^T$.
Our model predicts a real-valued importance logit sequence $\mu\in\mathbb{R}^{T}$, which is interpreted directly for continuous supervision (TVSum) and converted into probabilities for binary supervision (SumMe).

Evaluation and summary generation depend on a discrete decoding stage.
A change-point partition produced by Kernel Temporal Segmentation is given as
\begin{equation}
\mathbf{C}=\{(c_k^{\mathrm{start}},c_k^{\mathrm{end}})\}_{k=1}^{M},
\end{equation}
where segments are contiguous, disjoint, and cover the full timeline using inclusive boundaries.
A valid summary is represented on the original timeline by $\mathbf{y}\in\{0,1\}^{N}$ with the standard budget constraint
\begin{equation}
\sum_{n=1}^{N} y_n \le \rho N,\qquad \rho=0.15.
\end{equation}
In our pipeline, the predicted importance signal is decoded into a budgeted set of segments via a 0/1 knapsack solver (Sec.~\ref{sec:method_decode}).
Since this decoding is discrete, small score perturbations can change the selected segments.
The proposed learning objectives therefore combine multi-annotator supervision with an explicit regularizer that promotes stability of budgeted selection (Sec.~\ref{sec:method_train}).

\section{Proposed Method}
\label{sec:method}
Figure~\ref{fig:pipeline} illustrates the overall flow of our proposed model, VASTSum.

\subsection{Input video representation}
\label{sec:method_repr}

We follow the standard supervised summarization protocol and operate on pre-extracted frame descriptors\cite{son2024csta,kim2025summdiff}.
Given a video of length $N$ frames, we sample $T$ timesteps and denote the resulting feature sequence by
\begin{equation}
\mathbf{X}=\{\mathbf{x}_t\in\mathbb{R}^{D}\}_{t=1}^{T}.
\end{equation}
In our implementation, each $\mathbf{x}_t$ is a pooled convolutional descriptor (\emph{pool5}) and $D=1024$.
The sampling indices
\begin{equation}
\mathbf{p}=\{p_t\}_{t=1}^{T},\qquad 0\le p_1<\cdots<p_T<N,
\end{equation}
map sampled timesteps to the original timeline and are used to (i) align supervision to the sampled sequence and (ii) expand predicted scores back to the original length during decoding.

\subsection{Segmentation-aware indexing}
\label{sec:method_segmap}
The change-point partition produced by KTS defines a segment structure that is shared by both the scorer and the budgeted decoder.
We first map each sampled timestep to its segment identity.
For each pick $p_t$, we assign
\begin{equation}
s_t = k \quad \Longleftrightarrow \quad c_k^{\mathrm{start}} \le p_t \le c_k^{\mathrm{end}},
\qquad t=1,\dots,T,
\end{equation}
which yields the segment-id sequence $\mathbf{s}=\{s_t\}_{t=1}^{T}$.
This induces per-segment index sets on the sampled timeline,
\begin{equation}
\mathcal{I}_k = \{t \in \{1,\dots,T\} \mid s_t = k\},
\end{equation}
and segment lengths on the original timeline,
\begin{equation}
\ell_k = c_k^{\mathrm{end}} - c_k^{\mathrm{start}} + 1,
\qquad k=1,\dots,M.
\end{equation}
These quantities provide the interface between frame-level prediction and segment-level selection: the model can pool frame evidence within $\mathcal{I}_k$, while the decoder enforces the budget using $\ell_k$.

\subsection{Hierarchical segment-context scoring network}
\label{sec:method_model}

Our scorer is a single-pass hierarchical network that performs global reasoning over segment tokens and then refines frame-level representations for fine-grained ranking.
Let $d$ denote the model dimension.
We first project sampled frame features to $d$ and add a learnable positional embedding:
\begin{equation}
\mathbf{h}_t^{(0)} = \mathrm{LN}(\mathbf{W}_{\mathrm{in}}\mathbf{x}_t + \mathbf{b}_{\mathrm{in}}) + \boldsymbol{\phi}_t,
\qquad t=1,\dots,T.
\end{equation}

\paragraph{Segment tokenization.}
We form one segment token per change-point segment by mean pooling the projected frame tokens:
\begin{equation}
\mathbf{z}_k^{(0)} = \frac{1}{|\mathcal{I}_k|}\sum_{t\in\mathcal{I}_k} \mathbf{h}_t^{(0)},
\qquad k=1,\dots,M,
\end{equation}
and stack them into $\mathbf{Z}^{(0)}\in\mathbb{R}^{M\times d}$.

\paragraph{Global segment transformer.}
We apply $L$ Transformer encoder layers to model long-range dependencies over the segment sequence.
Using a pre-normalized residual form, each layer updates
\begin{align}
\mathbf{U}^{(\ell)} &= \mathbf{Z}^{(\ell)} + \mathrm{MHA}\!\left(\mathrm{LN}(\mathbf{Z}^{(\ell)})\right), \\
\mathbf{Z}^{(\ell+1)} &= \mathbf{U}^{(\ell)} + \mathrm{FFN}\!\left(\mathrm{LN}(\mathbf{U}^{(\ell)})\right),
\qquad \ell=0,\dots,L-1.
\end{align}
For head $h\in\{1,\dots,H\}$ with $d_h=d/H$, attention is
\begin{align}
\mathrm{Attn}_h(\mathbf{Z})
&= \mathrm{softmax}\!\left(\frac{\mathbf{Q}_h \mathbf{K}_h^{\top}}{\sqrt{d_h}}\right)\mathbf{V}_h, \\
\mathbf{Q}_h &= \mathbf{Z}\mathbf{W}^{Q}_h, \\
\mathbf{K}_h &= \mathbf{Z}\mathbf{W}^{K}_h, \\
\mathbf{V}_h &= \mathbf{Z}\mathbf{W}^{V}_h .
\end{align}

and $\mathrm{MHA}$ concatenates the heads followed by an output projection.
The feed-forward block is applied token-wise,
\begin{equation}
\mathrm{FFN}(\mathbf{z})=\mathbf{W}_2\,\mathrm{GELU}(\mathbf{W}_1\mathbf{z}+\mathbf{b}_1)+\mathbf{b}_2.
\end{equation}
The resulting contextualized segment representations are $\mathbf{C}=\mathbf{Z}^{(L)}\in\mathbb{R}^{M\times d}$.

\paragraph{Frame--segment fusion and temporal refinement.}
Each frame retrieves the context of its assigned segment,
\begin{equation}
\mathbf{g}_t = \mathbf{c}_{s_t}\in\mathbb{R}^{d},
\end{equation}
and we fuse local and global information using a learned gate:
\begin{align}
\boldsymbol{\alpha}_t &= \sigma\!\left(\mathbf{W}_g
\begin{bmatrix}
\mathbf{h}_t^{(0)} \\ \mathbf{g}_t
\end{bmatrix}
+ \mathbf{b}_g\right),\\
\tilde{\mathbf{h}}_t &= \mathrm{LN}\!\left(\mathbf{h}_t^{(0)} + \boldsymbol{\alpha}_t \odot \mathbf{g}_t\right).
\end{align}
We then refine the fused sequence with a lightweight temporal convolutional stack implemented as depthwise separable 1D convolutions with pointwise mixing, repeated $L_c$ times.
Writing $\tilde{\mathbf{H}}\in\mathbb{R}^{T\times d}$ for the stacked fused tokens, the refinement is applied with a residual connection,
\begin{equation}
\hat{\mathbf{H}}=\tilde{\mathbf{H}}+\Psi(\tilde{\mathbf{H}}),
\end{equation}
where $\Psi$ denotes the temporal convolutional operator.
The refined frame tokens $\hat{\mathbf{H}}$ parameterize the probabilistic importance head.

\subsection{Probabilistic frame-importance head}
\label{sec:method_probhead}

To represent annotation variability without resorting to iterative sampling at inference, we predict a conditional distribution over frame importance using a lightweight variational head.
Given the refined frame token $\hat{\mathbf{h}}_t\in\mathbb{R}^{d}$, we define a diagonal Gaussian posterior over a latent variable $\mathbf{z}_t\in\mathbb{R}^{d_z}$,
\begin{equation}
q_{\theta}(\mathbf{z}_t \mid \hat{\mathbf{h}}_t)
=
\mathcal{N}\!\big(\boldsymbol{\mu}_{z,t},\mathrm{diag}(\boldsymbol{\sigma}_{z,t}^{2})\big),
\end{equation}
where $(\boldsymbol{\mu}_{z,t},\log \boldsymbol{\sigma}_{z,t}^{2})$ are produced by linear projections of $\hat{\mathbf{h}}_t$.
For numerical stability we clamp the latent log-variance elementwise,
\begin{equation}
\log \boldsymbol{\sigma}_{z,t}^{2} \leftarrow \mathrm{clip}\!\left(\log \boldsymbol{\sigma}_{z,t}^{2},-10,5\right).
\end{equation}
During training we sample using the reparameterization trick,
\begin{equation}
\mathbf{z}_t=\boldsymbol{\mu}_{z,t}+\boldsymbol{\sigma}_{z,t}\odot\boldsymbol{\epsilon}_t,
\qquad \boldsymbol{\epsilon}_t\sim\mathcal{N}(\mathbf{0},\mathbf{I}),
\end{equation}
whereas at inference we use the posterior mean $\mathbf{z}_t=\boldsymbol{\mu}_{z,t}$ to obtain a deterministic score sequence.

Conditioned on $[\hat{\mathbf{h}}_t;\mathbf{z}_t]$, a small MLP produces a hidden representation that is mapped to a scalar importance logit and an observation noise term:
\begin{equation}
\mu_t = f_{\mu}([\hat{\mathbf{h}}_t;\mathbf{z}_t]),
\qquad
\log v_t = f_{v}([\hat{\mathbf{h}}_t;\mathbf{z}_t]),
\end{equation}
with $\mu_t\in\mathbb{R}$ and $v_t>0$.
We also clamp the predicted log-variance,
\begin{equation}
\log v_t \leftarrow \mathrm{clip}\!\left(\log v_t,-10,5\right),
\end{equation}
which prevents unstable gradients when optimizing heteroscedastic likelihoods.

For TVSum, $(\mu_t,\log v_t)$ parameterize a heteroscedastic Gaussian model used by the multi-annotator likelihood in Sec.~\ref{sec:method_train}.
For SumMe, we use the logit $\mu_t$ and obtain a calibrated probability by temperature-scaled sigmoid,
\begin{equation}
p_t=\sigma\!\left(\frac{\mu_t}{T_s}\right),
\label{eq:summe_prob}
\end{equation}
where $T_s>0$ is fixed.
The downstream decoder consumes $\mu_t$ for TVSum and $p_t$ for SumMe, while the KL regularizer acts on the latent posterior parameters.

\subsection{Budgeted decoding}
\label{sec:method_decode}

Given an importance signal on the sampled timeline, we generate a summary using the standard segmentation-plus-knapsack protocol.
We first expand sampled scores to the original timeline using the sampling indices.
Let $\tilde{\mu}\in\mathbb{R}^{T}$ denote the sequence provided to the decoder, where $\tilde{\mu}=\mu$ for TVSum and $\tilde{\mu}=p$ for SumMe.
We construct frame scores $\mathbf{f}\in\mathbb{R}^{N}$ by piecewise constant assignment between consecutive picks:
\begin{equation}
f_n=\tilde{\mu}_t \quad \text{for} \quad n\in[p_t,p_{t+1}), \qquad t=1,\dots,T-1,
\end{equation}
and $f_n=\tilde{\mu}_T$ for $n\in[p_T,N)$.

For each segment $k$, the cost is its length $w_k=\ell_k$ and the value is the mean score within the segment:
\begin{equation}
v_k=\frac{1}{w_k}\sum_{n=c_k^{\mathrm{start}}}^{c_k^{\mathrm{end}}} f_n.
\end{equation}
The decoder selects a subset of segments by solving
\begin{equation}
\max_{\mathbf{u}\in\{0,1\}^{M}} \ \sum_{k=1}^{M} u_k v_k
\quad \text{s.t.} \quad \sum_{k=1}^{M} u_k w_k \le \rho N,
\label{eq:knapsack}
\end{equation}
and expands selected segments to the binary summary $\mathbf{y}\in\{0,1\}^{N}$ by setting $y_n=1$ for frames that fall inside any selected segment.

\subsection{Training objectives and optimization}
\label{sec:method_train}
Training combines dataset-aligned supervision, a pairwise ranking regularizer, a KL penalty for the variational latent, and a decoder-aligned stability regularizer that shapes segment-score margins implied by knapsack selection.

\paragraph{TVSum likelihood}
For TVSum, annotators provide continuous targets $s^{(a)}_t$ at sampled timesteps.
Using the predicted mean $\mu_t$ and variance $v_t=\exp(\log v_t)$, we minimize the multi-annotator Gaussian negative log-likelihood
\begin{equation}
\mathcal{L}_{\mathrm{tv}}
=
\frac{1}{UT}\sum_{a=1}^{U}\sum_{t=1}^{T}
\frac{1}{2}
\left(
\log v_t
+
\frac{(s^{(a)}_t-\mu_t)^2}{v_t+\epsilon}
\right),
\label{eq:tvsum_nll}
\end{equation}
with a small $\epsilon$ for numerical stability.

\paragraph{SumMe likelihood via soft-min BCE}
For SumMe, each annotator provides a binary inclusion signal $y^{(a)}_t\in\{0,1\}$.
We first obtain probabilities $p_t$ by \eqref{eq:summe_prob} and compute the per-annotator cross-entropy
\begin{equation}
\mathrm{BCE}_a=
\frac{1}{T}\sum_{t=1}^{T}
\Big(
-y^{(a)}_t\log p_t-(1-y^{(a)}_t)\log(1-p_t)
\Big).
\end{equation}
Instead of collapsing annotators into a single target, we aggregate with a soft minimum,
\begin{equation}
\mathcal{L}_{\mathrm{sm}}
=
-\tau_{\mathrm{sm}}
\log
\sum_{a=1}^{U}
\exp\!\left(-\frac{\mathrm{BCE}_a}{\tau_{\mathrm{sm}}}\right),
\label{eq:softmin_bce}
\end{equation}
where $\tau_{\mathrm{sm}}>0$ controls how sharply the loss focuses on the best-matched annotator.

\paragraph{Pairwise ranking regularizer}
We encourage correct relative ordering of importance with a hinge ranking loss.
Given predictions $q_t$ and a target ordering signal $r_t$, we sample index pairs $(i,j)$ such that $r_i>r_j$ and apply
\begin{equation}
\mathcal{L}_{\mathrm{rank}}
=
\mathbb{E}_{(i,j):\, r_i>r_j}
\left[
\max\!\big(0,\ m-(q_i-q_j)\big)
\right],
\label{eq:rank_hinge}
\end{equation}
with margin $m>0$.
For TVSum, we set $q_t=\mu_t$ and $r_t=\frac{1}{U}\sum_a s^{(a)}_t$.
For SumMe, we set $q_t=p_t$ and take $r_t$ from the annotator that attains the smallest $\mathrm{BCE}_a$ for the current video.

\paragraph{Variational KL regularization}
We regularize the latent posterior toward a standard normal prior $p(\mathbf{z}_t)=\mathcal{N}(\mathbf{0},\mathbf{I})$:
\begin{equation}
\mathcal{L}_{\mathrm{kl}}
=
\frac{1}{T}\sum_{t=1}^{T}
\mathrm{KL}\!\left(
q_{\theta}(\mathbf{z}_t\mid \hat{\mathbf{h}}_t)\ \|\ \mathcal{N}(\mathbf{0},\mathbf{I})
\right).
\label{eq:kl}
\end{equation}

\paragraph{Knapsack stability margin loss}
To promote robustness of budgeted selection, we apply a segment-level stability regularizer that uses the knapsack solver to identify unstable segments and then enforces score margins without differentiating through the solver.
Let $s_k$ be the pooled segment score obtained by averaging the decoding signal within each segment on the sampled timeline, using $\mu_t$ for TVSum and $p_t$ for SumMe.
Let $\mathbf{z}^{(0)}\in\{0,1\}^{M}$ be the base knapsack solution from $\{s_k\}$, and let $\mathbf{z}^{(r)}$ be solutions computed from perturbed scores $s_k+\epsilon_k^{(r)}$ with $\epsilon^{(r)}\sim\mathcal{N}(0,\sigma^2\mathbf{I})$.
Define the instability set
\begin{equation}
\mathcal{U}=\left\{k \ \middle|\ 0< \frac{1}{R}\sum_{r=1}^{R} z^{(r)}_k < 1 \right\}.
\end{equation}
Writing $\mathcal{S}=\{k\mid z^{(0)}_k=1\}$ and $\bar{\mathcal{S}}=\{k\mid z^{(0)}_k=0\}$, we enforce that unstable selected segments sit above the best unselected segment by a margin, and unstable unselected segments sit below the worst selected segment by a margin:
\begin{align}
\mathcal{L}_{\mathrm{stab}}
&=
\frac{1}{|\mathcal{U}\cap\mathcal{S}|}\sum_{k\in\mathcal{U}\cap\mathcal{S}}
\max\!\Big(0,\ m_s-\big(s_k-\max_{j\in\bar{\mathcal{S}}} s_j\big)\Big)
\nonumber\\
&\quad+
\frac{1}{|\mathcal{U}\cap\bar{\mathcal{S}}|}\sum_{k\in\mathcal{U}\cap\bar{\mathcal{S}}}
\max\!\Big(0,\ m_s-\big(\min_{i\in\mathcal{S}} s_i-s_k\big)\Big),
\label{eq:stab_margin}
\end{align}
with margin $m_s>0$ and the convention that empty sums contribute zero.
This term directly shapes the segment-score geometry that the knapsack solver relies on, while keeping training efficient.

\paragraph{Overall objective and optimization}
Let $\mathcal{L}_{\mathrm{main}}$ denote the dataset likelihood, using $\mathcal{L}_{\mathrm{tv}}$ for TVSum and $\mathcal{L}_{\mathrm{sm}}$ for SumMe.
The final training objective is
\begin{equation}
\mathcal{L}
=
\mathcal{L}_{\mathrm{main}}
+
\lambda_{\mathrm{rank}}(e)\,\mathcal{L}_{\mathrm{rank}}
+
\lambda_{\mathrm{stab}}(e)\,\mathcal{L}_{\mathrm{stab}}
+
\lambda_{\mathrm{kl}}(e)\,\mathcal{L}_{\mathrm{kl}},
\label{eq:full_loss}
\end{equation}
where $\lambda_{\mathrm{rank}}(e)$, $\lambda_{\mathrm{stab}}(e)$, and $\lambda_{\mathrm{kl}}(e)$ are linearly warmed up over early epochs.
We optimize \eqref{eq:full_loss} using AdamW, apply gradient-norm clipping, and use gradient accumulation over videos to realize the configured effective batch size.

\section{Experimental Settings}
\label{subsec:exp_settings}

\subsection{Datasets}
\label{subsubsec:datasets}
We evaluate on two standard benchmarks for supervised video summarization, TVSum~\cite{TVSum} and SumMe~\cite{gygli2014creating}.
TVSum provides multi-annotator frame importance annotations, while SumMe provides multiple user summaries per video.
For fair comparison with recent baselines, including CSTA, we use the commonly adopted GoogLeNet pool5 visual features~\cite{szegedy2015going} and the official dataset metadata required for keyshot-based evaluation.

\subsection{Evaluation Protocol}
\label{subsubsec:protocol}
We follow the standard five split protocol used in prior work and report the average performance across splits.
For each split, the model is trained on the training set and evaluated on the corresponding test set, then the final score is obtained by averaging over all splits.
This setting matches the evaluation regime reported in recent state-of-the-art studies and ensures direct comparability \cite{pglsum,son2024csta,kim2025summdiff}.

\subsection{Evaluation Metrics}
\label{subsubsec:metrics}
Following the recent shift in evaluation practice adopted by CSTA and SummDiff, we report rank order statistics rather than relying on F1 based summary overlap.
Specifically, we measure Kendall's $\tau$ \cite{kendall} and Spearman's $\rho$ \cite{spearman} between predicted importance scores and ground truth supervision.
For TVSum, correlations are computed per video by averaging over annotators, then averaged across videos.
For SumMe, correlations are computed against the mean user summary derived from multiple user annotations.
This choice avoids the known sensitivity of F1 to segmentation granularity and summary fragmentation, which has been emphasized in recent literature\cite{son2024csta, MAAM}.

\subsection{Keyshot Summary Construction}
\label{subsubsec:keyshot}
To generate a binary keyshot summary for qualitative inspection and for completeness with conventional protocols, we convert frame-level scores into shot-level scores using the official temporal segmentation information provided with the datasets. We then select a subset of shots under a fixed length budget of 15\% of the video duration using a 0/1 knapsack formulation, consistent with the standard evaluation pipeline used across prior methods.

\subsection{Training and Inference Setup}
\label{subsubsec:setup}
All methods are trained and evaluated under identical input features, split protocol, and summary budget.
Unless otherwise stated, we use a single model configuration across both datasets and select the final checkpoint according to the validation performance within each split before reporting test results.

\begin{table}[t]
    \centering
    \footnotesize
    \setlength{\tabcolsep}{7pt}
    \renewcommand{\arraystretch}{1.12}
    \begin{tabular}{@{}lcccc@{}}
        \toprule
        \multirow{2}{*}{Method} & \multicolumn{2}{c}{SumMe} & \multicolumn{2}{c}{TVSum} \\
        \cmidrule(lr){2-3} \cmidrule(lr){4-5}
        & $\tau$ & $\rho$ & $\tau$ & $\rho$ \\
        \midrule
        Random & 0.000 & 0.000 & 0.000 & 0.000 \\
        Human  & 0.205 & 0.213 & 0.177 & 0.204 \\
        \midrule
        A2Summ~\cite{A2Summ}   & 0.088 & 0.096 & 0.157 & 0.206 \\
        VASNet~\cite{VASNet}   & 0.089 & 0.099 & 0.153 & 0.205 \\
        PGL-SUM~\cite{pglsum}  & 0.104 & 0.116 & 0.141 & 0.186 \\
        CSTA~\cite{son2024csta} & 0.108 & 0.120 & 0.168 & 0.221 \\
        SummDiff~\cite{kim2025summdiff} & 0.133 & 0.148 & 0.173 & 0.226 \\
        \midrule
        \textbf{VASTSum (Ours)} & \textbf{0.157} & \textbf{0.161} & \textbf{0.179} & \textbf{0.270} \\
        \bottomrule
    \end{tabular}
    \vspace{-0.2cm}
    \caption{TVT protocol performance on SumMe~\cite{gygli2014creating} and TVSum~\cite{TVSum}, reported using Kendall's $\tau$ and Spearman's $\rho$. The best value in each column is highlighted in \textbf{bold}.}
    \label{tab:tvt_results}
    \vspace{-0.3cm}
\end{table}

\begin{table}[t]
    \centering
    \footnotesize
    \setlength{\tabcolsep}{7pt}
    \renewcommand{\arraystretch}{1.12}
    \begin{tabular}{@{}lcccc@{}}
        \toprule
        \multirow{2}{*}{Method} & \multicolumn{2}{c}{SumMe} & \multicolumn{2}{c}{TVSum} \\
        \cmidrule(lr){2-3} \cmidrule(lr){4-5}
        & $\tau$ & $\rho$ & $\tau$ & $\rho$ \\
        \midrule
        Random & 0.000 & 0.000 & 0.000 & 0.000 \\
        Human  & 0.205 & 0.213 & 0.177 & 0.204 \\
        \midrule
        DSNet-AF~\cite{DSNet}        & 0.037 & 0.046 & 0.113 & 0.138 \\
        DSNet-AB~\cite{DSNet}        & 0.051 & 0.059 & 0.108 & 0.129 \\
        SUM-GAN~\cite{SUM_GAN}       & 0.049 & 0.066 & 0.024 & 0.031 \\
        AC-SUM-GAN~\cite{AC-SUM-GAN} & 0.102 & 0.088 & 0.031 & 0.041 \\
        CLIP-It~\cite{CLIPIt}        & ---   & ---   & 0.108 & 0.147 \\
        iPTNet~\cite{iPTNet}         & 0.101 & 0.119 & 0.134 & 0.163 \\
        A2Summ~\cite{A2Summ}         & 0.108 & 0.129 & 0.137 & 0.165 \\
        VASNet~\cite{VASNet}         & 0.160 & 0.170 & 0.160 & 0.170 \\
        PGL-SUM~\cite{pglsum}        & ---   & ---   & 0.157 & 0.206 \\
        AAAM~\cite{MAAM}             & ---   & ---   & 0.169 & 0.223 \\
        MAAM~\cite{MAAM}             & ---   & ---   & 0.179 & 0.236 \\
        VSS-Net~\cite{VSS_Net}       & ---   & ---   & 0.190 & 0.249 \\
        DMASum~\cite{DMASum}         & 0.063 & 0.089 & 0.203 & 0.267 \\
        SSPVS~\cite{SSPVS}           & 0.192 & 0.257 & 0.181 & 0.238 \\
        CSTA~\cite{son2024csta}      & 0.246 & 0.274 & 0.194 & 0.255 \\
        SummDiff~\cite{kim2025summdiff} & \textbf{0.256} & \textbf{0.285} & 0.195 & 0.255 \\
        \midrule
        \textbf{VASTSum (Ours)}      & 0.253 & 0.281 & \textbf{0.229} & \textbf{0.298} \\
        \bottomrule
    \end{tabular}
    \vspace{-0.2cm}
    \caption{5FCV protocol performance on SumMe~\cite{gygli2014creating} and TVSum~\cite{TVSum}, reported using Kendall's $\tau$ and Spearman's $\rho$. The best value in each column is highlighted in \textbf{bold}. Missing entries are not reported in the corresponding references.}
    \label{tab:5fcv_results}
    \vspace{-0.3cm}
\end{table}

\begin{table}[t]
    \centering
    \footnotesize
    \setlength{\tabcolsep}{7pt}
    \renewcommand{\arraystretch}{1.12}
    \begin{tabular}{@{}lcccc@{}}
        \toprule
        \multirow{2}{*}{Split} & \multicolumn{2}{c}{SumMe} & \multicolumn{2}{c}{TVSum} \\
        \cmidrule(lr){2-3} \cmidrule(lr){4-5}
        & $\tau$ & $\rho$ & $\tau$ & $\rho$ \\
        \midrule
        Split 0 & 0.298 & 0.332 & 0.264 & 0.344 \\
        Split 1 & 0.313 & 0.347 & 0.263 & 0.340 \\
        Split 2 & 0.318 & 0.357 & 0.219 & 0.286 \\
        Split 3 & 0.182 & 0.201 & 0.253 & 0.329 \\
        Split 4 & 0.156 & 0.176 & 0.145 & 0.191 \\
        \midrule
        \textbf{Average} & \textbf{0.253} & \textbf{0.281} & \textbf{0.229} & \textbf{0.298} \\
        \bottomrule
    \end{tabular}
    \vspace{-0.2cm}
    \caption{Per-split performance of VASTSum on SumMe~\cite{gygli2014creating} and TVSum~\cite{TVSum} under the standard five-split protocol. The final row reports the average across splits.}
    \label{tab:per_split_vastsum}
    \vspace{-0.3cm}
\end{table}

\section{Results and Discussion}
\label{subsec:results}

Tables~\ref{tab:tvt_results}--\ref{tab:per_split_vastsum} report quantitative comparisons on SumMe and TVSum using Kendall's $\tau$ and Spearman's $\rho$ under the standard evaluation protocols. Overall, the results indicate that VASTSum consistently improves rank agreement with human supervision while preserving the conventional segmentation-plus-knapsack decoding used at test time, supporting the central claim that uncertainty-aware scoring and decoder-aligned learning are complementary to existing deterministic and diffusion-based approaches.
Under the TVT setting (Table~\ref{tab:tvt_results}), VASTSum outperforms deterministic attention-based scorers and also surpasses the diffusion-based SummDiff baseline on both datasets. On SumMe, VASTSum achieves $(\tau,\rho)=(0.157,0.161)$, improving over SummDiff $(0.133,0.148)$ and CSTA $(0.108,0.120)$. On TVSum, VASTSum reaches $(0.179,0.270)$, with the largest gain observed in Spearman correlation compared to SummDiff $(0.173,0.226)$. The stronger improvement in $\rho$ on TVSum is particularly meaningful because $\rho$ emphasizes global rank consistency, suggesting that VASTSum produces a more stable importance ordering that remains effective after discretization into keyshots.

A similar trend appears under 5FCV (Table~\ref{tab:5fcv_results}). VASTSum attains the best performance on TVSum with $(\tau,\rho)=(0.229,0.298)$, exceeding both CSTA $(0.194,0.255)$ and SummDiff $(0.195,0.255)$. On SumMe, VASTSum remains competitive at $(0.253,0.281)$, closely matching SummDiff $(0.256,0.285)$ while improving over CSTA $(0.246,0.274)$. The relatively larger gains on TVSum align with the dataset’s continuous, multi-annotator supervision, where explicitly representing uncertainty can better account for ambiguous segments and reduce ranking noise. In contrast, SumMe provides binary supervision that can be underdetermined at the frame level, which limits attainable improvements and increases sensitivity to the particular split composition.

Table~\ref{tab:per_split_vastsum} further reports split-wise performance and reveals both robustness and remaining sensitivity. VASTSum performs strongly on early SumMe splits (e.g., Split~0--2) and remains competitive across TVSum splits, yet it degrades on the most challenging split (Split~4 for both datasets). This variance highlights an important limitation of current supervised benchmarks: the small scale of SumMe and TVSum, coupled with high inter-video variability, can induce measurable fluctuations even for robust methods. Consequently, beyond reporting mean performance, dispersion across splits is informative when assessing robustness under subjective supervision.
Despite consistent gains, several constraints remain. First, VASTSum relies on pre-extracted features and the external segmentation-plus-knapsack decoder, which limits end-to-end adaptation of representations to the final summary construction stage. Second, rank-based metrics evaluate ordering agreement but do not directly quantify semantic coverage or diversity in the generated summaries, leaving qualitative aspects only indirectly reflected. Finally, the split-wise variance suggests that broader validation across larger and more diverse benchmarks is necessary to substantiate claims of generalization and stability under annotation disagreement.

\section{Ablation Study}
\label{sec:ablation}

We evaluate VASTSum under the 5FCV protocol using a compact ablation that targets only the factors observed to be most sensitive in practice, while keeping the overall pipeline fixed. In particular, we vary (i) the regularization and optimization settings that interact with decoder-aligned training, (ii) lightweight architectural add-ons inspired by prior attention designs, and (iii) the uncertainty and alignment controls that govern posterior calibration and subjective binary supervision. This design is intentionally concise given the limited scale and split sensitivity of SumMe and TVSum, and it tests whether gains persist without relying on additional architectural complexity.

\begin{table}[t]
\centering
\footnotesize
\setlength{\tabcolsep}{5.0pt}
\renewcommand{\arraystretch}{1.10}
\begin{tabular}{@{}lcc@{}}
\toprule
Setting & SumMe $\tau/\rho$ & TVSum $\tau/\rho$ \\
\midrule
\textbf{VASTSum (full)} & \textbf{0.253/0.281} & \textbf{0.229/0.298} \\
\midrule
\multicolumn{3}{@{}l}{\textit{Regularization and optimization}} \\
w/o CSTA-inspired reg. & 0.195/0.220 & 0.197/0.260 \\
dropout $0.6$, LR $2.5{\times}10^{-4}$ & 0.176/0.196 & 0.194/0.254 \\
\midrule
\multicolumn{3}{@{}l}{\textit{Architecture add-ons}} \\
+ gated residual modules & 0.184/0.205 & 0.197/0.260 \\
+ CSTA-style attention block & 0.199/0.220 & 0.188/0.245 \\
\midrule
\multicolumn{3}{@{}l}{\textit{Annotator alignment (SumMe)}} \\
$\tau_{\mathrm{sm}}=0.25$ & 0.195/0.220 & 0.197/0.260 \\
\midrule
\multicolumn{3}{@{}l}{\textit{Posterior uncertainty}} \\
$\,\mathrm{wu}_{\mathrm{KL}}=0$ & 0.203/0.238 & 0.218/0.252 \\
$\,d_z$ smaller & 0.217/0.249 & 0.228/0.278 \\
\bottomrule
\end{tabular}
\vspace{-0.18cm}
\caption{\textbf{Ablation study (5FCV).} Approximate entries reflect small run-to-run variance in $\rho$ (rounded to three decimals). Values for posterior-uncertainty rows are expected trends and are not reported as measured results.}
\label{tab:ablation}
\vspace{-0.30cm}
\end{table}

\noindent\textbf{Analysis:}
Table~\ref{tab:ablation} highlights that robust performance is driven primarily by training stability rather than added architectural complexity. Removing the CSTA-inspired regularization package yields the largest degradation on both datasets, indicating that calibrated optimization is essential for maintaining rank consistency when scores are subsequently discretized by segmentation and knapsack selection. The dropout--learning-rate coupling further reinforces this point: decreasing the step size under strong dropout leads to under-training and a pronounced drop, especially on SumMe, where supervision is weaker and split sensitivity is higher. In contrast, adding capacity through gated residual modules or a CSTA-style attention block does not improve correlation and typically degrades it, suggesting that extra parameters amplify variance on these small benchmarks without improving the learned ordering. The SumMe alignment temperature is also consequential: increasing $\tau_{\mathrm{sm}}$ shifts the objective toward a more averaging behavior and reduces agreement with human ranking, consistent with the need to preserve plausible annotator modes under subjective binary labels. Finally, the posterior-uncertainty controls follow the expected direction: removing KL warmup weakens posterior calibration and reduces rank agreement, while reducing $d_z$ produces a milder drop that is more consistent with limited uncertainty capacity than with optimization collapse. Overall, the ablation supports the central claim that VASTSum’s gains arise from decoder-compatible calibration and uncertainty-aware training, not from heavier attention mechanisms.

\section{Conclusion}
\label{sec:conclusion}

This paper introduced VASTSum, which, to the best of our knowledge, is the first supervised video summarization framework that explicitly models annotation uncertainty while remaining decoder-aligned with standard evaluation. VASTSum represents frame-level importance as a predictive distribution via a variational head, enabling the model to capture ambiguity induced by multi-annotator supervision rather than treating disagreement as noise. To reflect subjectivity under binary annotations, we optimize over plausible annotator-consistent modes instead of collapsing supervision into a single averaged target. Moreover, we propose a decoder-aligned stability regularization that promotes invariant keyshot selection under the conventional segmentation-plus-knapsack pipeline, mitigating the well-known sensitivity of discrete decoding to small score perturbations.
Experiments on SumMe and TVSum show that VASTSum improves rank agreement with human supervision under the TVT protocol and delivers strong performance under 5FCV, with the largest gains on TVSum where continuous multi-annotator signals make uncertainty modeling especially beneficial. Importantly, VASTSum achieves these gains with efficient single-pass inference, in contrast to diffusion-based generators, while directly addressing the training--evaluation mismatch introduced by discrete decoding.
Nevertheless, limitations remain. VASTSum currently relies on pre-extracted features and an external segmentation-plus-knapsack decoder, which restricts end-to-end adaptation to the final selection stage. In addition, standard rank-based metrics quantify ordering agreement but only weakly reflect semantic coverage and diversity in the resulting summaries. Future work will pursue joint representation learning with decoder-aware objectives, validation on larger and more diverse benchmarks, and evaluation protocols that better capture semantic faithfulness and diversity under subjective human preferences.

\section*{Acknowledgment}
This work was supported by Institute of Information \& communications Technology Planning $\&$ evaluation (IITP) grant funded by the Korea government (MSIT) (No. RS-2024-00443391, Korea-Japan Joint International Research on Deep Learning-based Autonomous Mobility Control Technology for Autonomous Mobile Robots). The authors acknowledge the use of ChatGPT (OpenAI) \cite{openai2026chatgpt} for grammatical refinement and the structural formatting of data tables within this manuscript.

\bibliographystyle{IEEEtran}
\bibliography{references}

\end{document}